%% file: neurips_2025.tex
\documentclass{article}

    \usepackage[final]{neurips_2025}

\usepackage{amsmath}
\usepackage[ruled,vlined]{algorithm2e}
\usepackage{graphicx}
\usepackage{multirow}
\usepackage{makecell}
\usepackage{booktabs}
\usepackage{xcolor}
\usepackage{wrapfig}

\usepackage[utf8]{inputenc} 
\usepackage[T1]{fontenc}    
\usepackage{hyperref}       
\usepackage{url}            
\usepackage{booktabs}       
\usepackage{amsfonts}       
\usepackage{nicefrac}       
\usepackage{microtype}      
\usepackage{xcolor}         

\usepackage{hyperref}
\hypersetup{
    colorlinks=true,
    linkcolor=blue,
    filecolor=magenta,
    urlcolor=cyan,
}

\title{LIMOPro: Reasoning Refinement for Efficient and Effective Test-time Scaling}

\author{Yang Xiao$^{1}$ \quad Jiashuo Wang$^{1}$ \quad  Ruifeng Yuan$^{1}$  \quad \textbf{Chunpu Xu}$^{1}$ \\
\textbf{Kaishuai Xu}$^{1}$ \quad \textbf{Wenjie Li}$^{1}$\footnotemark[2] \quad \textbf{Pengfei Liu}$^{2,3}$\footnotemark[2] \\
$^{1}$The Hong Kong Polytechnic University \quad $^{2}$Shanghai Jiao Tong University \quad $^{3}$SII \\
\texttt{yang-alan.xiao@connect.polyu.hk} \quad \texttt{csjwang@comp.polyu.edu.hk} 
}

\begin{document}

\maketitle
{
\renewcommand{\thefootnote}{\fnsymbol{footnote}}
\footnotetext[2]{Corresponding authors.}
}

\input{chapter/abstract}

\input{chapter/intro}

\input{chapter/beyondLimo}

\input{chapter/experiment}

\input{chapter/related_work}

\newpage
\input{chapter/acknowledgement}

\bibliography{ref}
\bibliographystyle{plain}

\newpage
\appendix

\input{chapter/append}


\newpage
\input{chapter/checklist}

\end{document}

%% file: chapter/abstract.tex
\begin{abstract}


Large language models (LLMs) have demonstrated remarkable reasoning capabilities through test-time scaling approaches, particularly when fine-tuned with chain-of-thought (CoT) data distilled from more powerful large reasoning models (LRMs). However, these reasoning chains often contain verbose elements that mirror human problem-solving, categorized as progressive reasoning (the essential solution development path) and functional elements (verification processes, alternative solution approaches, and error corrections). While progressive reasoning is crucial, the functional elements significantly increase computational demands during test-time inference. We introduce \textbf{PIR (Perplexity-based Importance Refinement)}, a principled framework that quantitatively evaluates the importance of each reasoning step based on its impact on answer prediction confidence. PIR systematically identifies and selectively prunes only low-importance functional steps while preserving progressive reasoning components, creating optimized training data that maintains the integrity of the core solution path while reducing verbosity. Models fine-tuned on PIR-optimized data exhibit superior test-time scaling properties, generating more concise reasoning chains while achieving improved accuracy (+0.9\% to +6.6\%) with significantly reduced token usage (-3\% to -41\%) across challenging reasoning benchmarks (AIME, AMC, and GPQA Diamond). Our approach demonstrates strong generalizability across different model sizes, data sources, and token budgets, offering a practical solution for deploying reasoning-capable LLMs in scenarios where efficient test-time scaling, response time, and computational efficiency are valuable constraints. Code and dataset are available at the \href{https://github.com/GAIR-NLP/LIMOPro.git}{LIMOPro}.

\begin{figure}[ht]
\vspace{-10pt}
    \centering
    \includegraphics[width=1\linewidth]{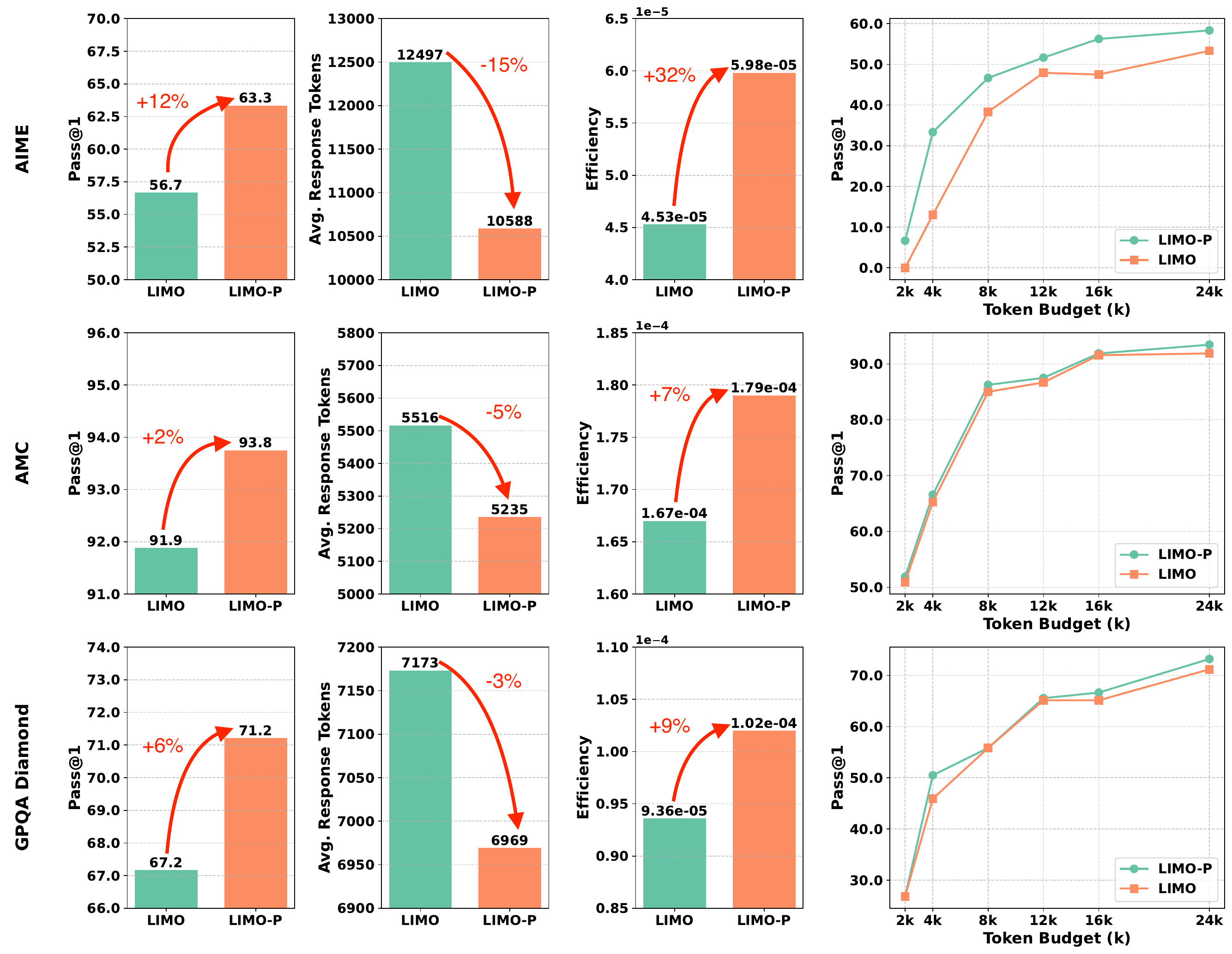}
    \caption{Our PIR framework (implemented as LIMO-P) optimizes the efficiency-effectiveness tradeoff in LLM reasoning across AIME, AMC, and GPQA Diamond benchmarks: it consistently enhances accuracy while concurrently reducing response tokens, thus improving computational efficiency, demonstrating that selectively pruning low-importance functional steps produces more concise, faster, and more accurate reasoning chains.}
    \label{fig:cover}
   \vspace{-20pt}
\end{figure}
\end{abstract}

%% file: chapter/intro.tex
\section{Introduction}
\vspace{-10pt}
Large language models (LLMs) have demonstrated remarkable capabilities in complex reasoning tasks through chain-of-thought (CoT) \citep{wei2022chain}, where models generate step-by-step solutions to problems. Recent advances of test-time scaling \citep{jaech2024openai,snell2024scaling} can significantly enhance LLMs' reasoning abilities by increasing the compute at test time. One approach to the test-time scaling involves fine-tuning LLMs on high-quality reasoning data distilled from more powerful large reasoning models (LRMs) \citep{muennighoff2025s1,ye2025limo}. LRMs like DeepSeek-R1 \citep{guo2025deepseek}, OpenAI o1 \citep{jaech2024openai}, and QwQ \citep{qwq32b} represent the state of the art in this paradigm, producing reasoning chains that lead to accurate solutions.

However, this approach faces a significant challenge: reasoning chains distilled from LRMs often contain numerous functional elements that, while reflecting human problem-solving processes, possibly produce unnecessarily verbose outputs \citep{fan2025missing, wang2025thoughts, chen2024not}. In a typical mathematical scenario, an LRM might solve a problem by establishing an initial solution path, verifying calculations, identifying errors, revising the approach, and ultimately confirming the final answer. This thorough process generates lengthy reasoning chains with redundant or marginally valuable steps. When these verbose chains are used to fine-tune target models, the resulting models inevitably adopt similar behaviors, producing equally elaborate reasoning sequences despite many steps contributing minimally to solution accuracy. Consequently, inference time increases substantially, along with computational demands and response latency. This inefficiency poses a considerable obstacle to implementing reasoning-enhanced LLMs in practical applications where timely, precise responses are essential.

To address this challenge, we introduce \textbf{PIR (Perplexity-based Importance Refinement)}, a novel framework that systematically refines reasoning chains to optimize the efficiency-effectiveness balance. Our refinement approach builds upon three key innovations that work in concert: (1) We develop a systematic methodology to classify functional patterns in complex reasoning chains, identifying four distinct modes—progressive reasoning and three types of functional steps: verification, multi-method validation, and error correction. (2) Through comprehensive analysis across diverse problem domains, our analysis indicates that progressive reasoning forms the essential logical backbone, directly advancing solution derivation, and must be preserved intact. In contrast, functional steps frequently introduce computational overhead with redundancies that can be strategically pruned without compromising solution integrity. This differential treatment—preserving progressive reasoning while selectively optimizing functional elements—maintains the core problem-solving logic while significantly improving computational efficiency.
(3) Building on (1) and (2), we propose the PIR metric, which quantitatively measures each functional step's contribution to the final solution by comparing answer perplexity with and without specific steps. This perplexity-based evaluation provides a principled mechanism to refine reasoning chains by identifying and selectively removing low-importance functional steps while preserving the progressive reasoning chain. By selectively targeting only non-essential functional components, our refinement approach maintains the logical coherence of the solution process while significantly reducing verbosity.

By applying our refinement framework to datasets distilled from different foundation models (LIMO \citep{ye2025limo} from Deepseek-R1, S1 \citep{muennighoff2025s1} from Gemini Flash Thinking \citep{googleThinkingGenerative}, and LIMO-V2 \citep{ye2025limo} from QwQ), we create PIR-optimized training datasets. Models fine-tuned on these refined datasets maintain or enhance accuracy while significantly reducing response length compared to models trained on the original unrefined data. Our experiments across challenging reasoning benchmarks demonstrate that PIR-refined models consistently outperform their counterparts in both effectiveness and efficiency, achieving up to 71\% efficiency improvement.

The contributions of our work include: 1. A novel perplexity-based refinement framework (PIR) for quantifying the importance of reasoning steps and optimizing reasoning chains, balancing efficiency and effectiveness. 2. A systematic analysis of reasoning patterns in reasoning problem-solving, providing insights into the structure and function of different reasoning elements. 3. Comprehensive empirical validation showing that PIR-refined models achieve improved accuracy (+0.9\% to +6.6\%) with significantly reduced token usage (-3\% to -41\%) across diverse benchmarks. 4. Demonstration of the framework's generalizability across different model sizes, data sources, and token budgets.

Our work addresses a critical gap in current approaches to LLM reasoning enhancement, offering a practical solution for more efficient reasoning without sacrificing solution quality. By systematically refining training data to preserve essential reasoning while eliminating redundant functional steps (verification, validation, and error correction processes), we enable LLMs to produce concise yet equally effective reasoning chains, advancing the practical deployment of reasoning-capable LLMs across scenarios where response time and computational resources are valuable constraints.

%% file: chapter/beyondLimo.tex
\section{Reasoning Refinement: Reasoning Optimization Framework}
Reasoning chains produced by LRMs typically contain numerous functional steps—including verification processes, multiple solution approaches, and error corrections—that mirror human problem-solving but significantly increase computational overhead without proportionally enhancing solution accuracy. This section presents our Perplexity-based Importance Refinement (PIR) framework, which quantitatively evaluates reasoning step importance and systematically refines reasoning chains by preserving essential solution elements while removing less valuable components. 


\subsection{Problem Formulation}

In this paper, we address the challenge of optimizing reasoning chains for complex reasoning tasks. Formally, we consider a dataset $\mathcal{D}$ containing question-reasoning-answer triplets $(q, r, a)$, where $q \in \mathcal{Q}$ represents a reasoning problem, $r \in \mathcal{R}$ is the reasoning chain, and $a \in \mathcal{A}$ is the answer. We define a reasoning chain $r$ as a sequence of intermediate steps $\{s_1, s_2, ..., s_n\}$, where each step $s_i$ represents a logical deduction that bridges the gap between the question and the final answer.

Our goal is to refine each reasoning chain $r$ into an optimized version $r'$ such that: (1) The answer accuracy is preserved: $f(q, r') = f(q, r) = a$;
(2) The token length is reduced: $|r'| < |r|$;
(3) The essential reasoning logic is maintained without harming the quality of the dataset.


\subsection{Theoretical Foundations}

\paragraph{Cognitive Reasoning Patterns}
Through extensive analysis of reasoning patterns, we identify four representative distinct modes \citep{gandhi2025cognitive,fu2024efficiently} that characterize problem-solving processes: \textbf{(1) Progressive Reasoning}, characterized by forward-chaining inference that follows a deductive logical progression from premises to conclusion, forming the essential backbone of solution development; \textbf{(2) Verification}, which represents metacognitive monitoring processes where previous calculations are systematically validated for accuracy, often using phrases like ``Let me check''; \textbf{(3) Multi-method Validation}, demonstrating convergent thinking by applying diverse methodological approaches to reinforce conclusions, potentially introducing redundancy; and \textbf{(4) Error Correction} embodies a self-regulatory mechanism through which logical inconsistencies, computational errors, or potential mistakes are identified and remediated. This pattern captures the process of recognizing when a path of reasoning may be flawed, reassessing assumptions, and correcting course. While progressive reasoning constitutes the critical path to solution derivation, the other three functional patterns, though valuable in human-like reasoning, often contain redundancies that can be optimized without compromising solution integrity. Details about the four reasoning patterns and corresponding cases can be found in  Appendix \ref{append:pattern case}. 

\paragraph{PIR: Perplexity-Based Importance Refinement of Reasoning Steps}
PIR quantifies reasoning step importance by measuring perplexity changes when specific steps are removed. The indicator compares answer perplexity with and without a particular reasoning step:



\begin{equation}
\text{PIR}_{\theta}(x_i|x_{1:n}) = \log\left(\frac{\text{PPL}_{\theta}(R \setminus \{x_i\})}{\text{PPL}_{\theta}(R)}\right)
\end{equation}

Where $\text{PPL}_{\theta}(R)$ and $\text{PPL}_{\theta}(R \setminus \{x_i\})$ represent perplexities calculated by model $\theta$ with and without step $i$:
\begin{align}
\text{PPL}_{\theta}(R) &= \exp\left(-\frac{1}{m} \sum_{j=1}^{m} \log p_{\theta}(a_j|x_{1:n}, a_{<j})\right) \\
\text{PPL}_{\theta}(R \setminus \{x_i\}) &= \exp\left(-\frac{1}{m} \sum_{j=1}^{m} \log p_{\theta}(a_j|x_{1:i-1}, x_{i+1:n}, a_{<j})\right)
\end{align}

Here, $m$ represents the number of answer tokens, $a_j$ is the $j$-th answer token, $x_{1:n}$ represents all reasoning steps, and $x_{1:i-1}, x_{i+1:n}$ represents all steps except step $i$. The perplexity calculations are performed using a model $\theta$. A higher PIR value indicates greater step importance: when a critical reasoning step is eliminated, the model becomes significantly less confident in generating the correct answer, lacking essential information for solution derivation, resulting in higher perplexity and thus a higher PIR score.

\begin{figure}[t]
    \centering
    \includegraphics[width=0.9\linewidth]{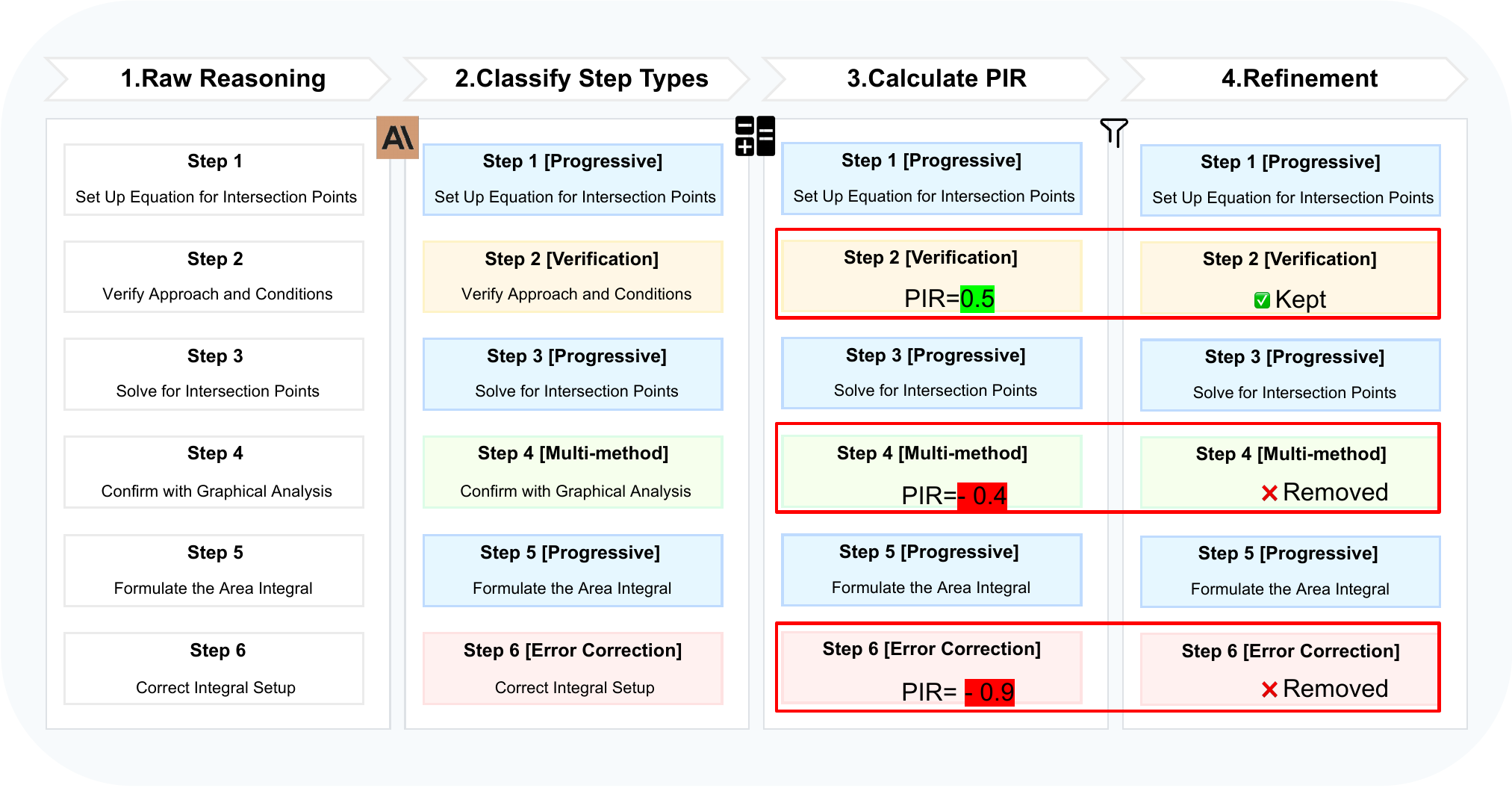}
    \caption{PIR framework pipeline for reasoning optimization: raw reasoning is segmented into logical steps, step is classified into reasoning patterns, PIR value is calculated to quantify step importance, and low-value functional steps are filtered while preserving progressive reasoning, resulting in more efficient reasoning chains.}
    \label{fig:pipeline}
    \vspace{-15pt}
\end{figure}


\subsection{Analysis and Optimization of Reasoning Chains}
Our approach first employs a hierarchical decomposition process where reasoning chains are segmented into logical steps by Claude 3.7 Sonnet \citep{anthropicClaudeSonnet}, with each step typically comprising multiple coherent sentences that form a cohesive reasoning unit. For classification, we implement a two-phase system: initially, a rule-based pattern matching component identifies steps containing characteristic linguistic markers (such as ``Let me check'' for verification or ``I made a mistake'' for error correction). For steps lacking explicit markers, we again apply Claude 3.7 Sonnet to perform contextual analysis, capturing more nuanced reasoning structures and assigning appropriate pattern classifications.

To validate our classification methodology, we randomly select 5\% of classified steps across our datasets for human evaluation. Four postgraduates independently assess whether each step is correctly classified according to our defined patterns. Steps are considered correctly classified only when all four annotators unanimously agree with the system's classification. This rigorous evaluation protocol revealed that 93.4\% of the steps are unanimously verified as correctly classified, demonstrating the robust performance of our hybrid classification approach for reasoning step categorization.

Building on this classification system and the PIR metric, we implemented a targeted pruning approach that selectively identifies and removes low-importance functional steps while maintaining data quality and preserving reasoning integrity. This process is shown in Figure \ref{fig:pipeline}. Unlike approaches that might indiscriminately compress reasoning chains, our method specifically targets functional steps (verification, multi-method validation, and error correction) while preserving all progressive reasoning steps, which constitute the essential deductive core of the solution process. For each identified functional step, we compute its PIR value to quantify its importance based on the impact its removal has on answer prediction confidence. The process then selectively removes functional steps with the lowest PIR values according to a predefined ratio threshold, resulting in more concise reasoning chains that maintain effectiveness while significantly reducing verbosity.

This selective pruning approach creates more efficient reasoning chains that maintain the integrity of core problem-solving logic while eliminating redundant verification processes. The resulting optimized chains are both more efficient to process and more effective as training exemplars for downstream models, without sacrificing the quality and completeness of the essential reasoning. The Algorithm \ref{alg:alg1} in the appendix demonstrates this whole process. In our experiments, we apply Qwen2.5-32B-Instruct to calculate PPL.

\begin{table}[t]
\vspace{-16pt}   
    \centering
    \small 
    \caption{Dataset Statistics. For each dataset, we report the data source, number of samples, total token count, and the distribution of tokens across four distinct reasoning patterns. }
    \label{tab:dataset_stats}
    \resizebox{\textwidth}{!}{%
    \begin{tabular}{lccccccc}
        \toprule
        \textbf{Dataset} & \textbf{Source} & \textbf{Samples} & \textbf{Tokens} &\textbf{\makecell{Progressive\\Reasoning}}&\textbf{Verification}&\textbf{\makecell{Multi-method\\Validation}}& \textbf{\makecell{Error\\Correction}}\\
        \midrule
        S1K & Gemini & 1,000 & 4,509,505&71.4\%&9.2\%&13.9\%&5.5\% \\
        LIMO & DeepSeek-R1 & 817 & 5,144,004 &59.7\%&11.8\%&10.9\%&17.6\% \\
        LIMO-V2 & QwQ & 800 & 8,866,950&64.3\%&9.8\%&12.7\%&13.2\% \\
        \bottomrule
    \end{tabular}
    }
    \vspace{-15pt}  
\end{table}

%% file: chapter/experiment.tex
\newcommand{\annotate}[3]{%
    #1\raisebox{-0.5ex}{\scriptsize\textcolor{#2}{#3}}%
}

\section{Experimental Results and Analysis}

\subsection{Experimental Setup}

\paragraph{Training Dataset}

To empirically validate our proposed PIR reasoning refinement framework, we conduct experiments using established datasets from prior work. As demonstrated in Table \ref{tab:dataset_stats}, our approach leverages three representative reasoning datasets: LIMO (distilled from DeepSeek R1), LIMO-V2 (distilled from QwQ), and S1K (distilled from Gemini Thinking). These datasets, distilled from different foundation models, represent diverse reasoning patterns and problem-solving approaches. Applying our framework across these varied sources allows us to validate the generalizability of our method across different data sources, ensuring that the PIR optimization is robust and not dependent on specific characteristics of any single source model. We apply our PIR framework to these datasets to create optimized versions: LIMO-P, LIMO-V2-P, and S1K-P. Additionally, we construct several variant datasets by implementing different refinement ratios to investigate the optimal balance between conciseness and effectiveness. Detailed information regarding different refinement ratios and the specific token counts of the optimized datasets is provided in the Appendix \ref{append:full_data_statistic}.

\paragraph{Benchmark Datasets}

To rigorously assess our methodology, we utilized three representivate reasoning-intensive benchmarks: \textbf{(1)} \textbf{AIME24} evaluation set encompasses 30 challenging problems from the American Invitational Mathematics Examination administered in early 2024, requiring sophisticated reasoning across mathematical domains; \textbf{(2)} \textbf{GPQA Diamond} corpus incorporates 198 doctoral-level scientific inquiries across biological, chemical, and physical disciplines, presenting formidable challenges that even subject matter experts struggle to master fully; \textbf{(3)} \textbf{AMC23} includs 40 problems from the AIMO progress prize competition. Using data after 2023.


\paragraph{Performance metrics: Reasoning Effectiveness and Efficiency}
To evaluate reasoning \textbf{effectiveness ($\text{ACC}$)}, we employ the pass@1 accuracy as our primary performance indicator. For each problem in our benchmark, we sample eight responses from the model and calculate $\text{ACC}$ under the Zero-shot Chain-of-Thought (CoT) setting with the instruction of: ``Please reason step by step, and put your final answer within boxed.'' We utilize Qwen2.5-Math evaluators \citep{yang2024qwen2} to systematically assess solution correctness across all solutions, with each sampling conducted at a temperature setting of 0.7 to balance deterministic reasoning with exploration of solution paths. Building on our effectiveness measure, we quantify reasoning \textbf{efficiency ($\text{EFF}$)} as the ratio between model performance and resource utilization: $\text{EFF} = \text{ACC}/\text{TOK}$, where $\text{TOK}$ represents the average number of response tokens across all benchmark problems. This efficiency metric captures the utility produced per unit of test-time resource consumption, effectively highlighting the critical trade-off between performance and computational cost. 

\paragraph{Evaluation of PIR Framework Across Multiple Training Datasets} We establish baseline models (LIMO, LIMO-V2, and S1-32B) that were trained on the original unmodified datasets using Qwen2.5-32B-Instruct \citep{yang2024qwen2}. These baseline models were obtained from their official Hugging Face repositories.\footnote{https://huggingface.co/} To test our PIR framework, we fine-tune the same Qwen2.5-32B-Instruct base model on the pruned datasets: LIMO-P, LIMO-V2-P, and S1K-P, using identical training scripts as described in the original papers and yielding our PIR-optimized models: LIMO-P, LIMO-V2-P, and S1-32B-P. All evaluations follow consistent testing protocols across methods to maintain comparative validity.

\paragraph{Comparison Against Alternative Reasoning Optimization Methods} We compare our PIR framework with other competitive reasoning optimization methods. First, we establish \textbf{S1-32B} as our primary baseline trained on the original unmodified S1K dataset without any reasoning optimization. We then compare against two leading competing methods from prior work: Prompt Reduction \citep{ding2024break}, denoted as \textbf{S1-PROMPT}, which develops innovative prompting strategies
that encourage LLMs to use shortcuts to quickly exploit reasoning clues and bypass detailed procedural steps. We apply this method for model S1-32B as a baseline; and SPIRIT \citep{cui2025stepwise}, denoted as \textbf{S1-SPIRIT}, which applies a non-discriminative pruning approach for all reasoning steps based solely on perplexity values. We apply this method to the S1K dataset to get a pruned dataset, and fine-tuning Qwen2.5-32B-Instruct with the pruned dataset. Unlike our PIR method that selectively preserves essential progressive reasoning steps, S1-SPIRIT's uniform filtering risks removing critical components necessary for accurate solutions. Additionally, we include an ablation of our method, Rule-based Filtering (denoted as \textbf{S1-RULE}), which identifies functional step categories but randomly removes steps from these categories without using PIR metrics to determine their importance.\footnote{We specifically chose S1K dataset for these comparisons because competing methods like SPIRIT require perplexity calculations for all sentences, which would be computationally prohibitive for the significantly larger LIMO or LIMO-V2 datasets, resulting in excessive computational costs and processing time.}

\begin{table}[t]
\centering
\setlength{\tabcolsep}{1.8pt}
\renewcommand{\arraystretch}{1.25}
\caption{Experimental results comparing baseline models with their PIR-optimized variants (-P) across reasoning benchmarks. Metrics include accuracy (ACC), token length (TOK), and efficiency (EFF). }
\label{main-exp}
\resizebox{\textwidth}{!}{%
\begin{tabular}{clllllllll}
\toprule
\multirow{2}{*}{\textbf{Model}} & \multicolumn{3}{c}{\textbf{AIME}} & \multicolumn{3}{c}{\textbf{AMC}} & \multicolumn{3}{c}{\textbf{GPQA Diamond}}  \\
\cmidrule(lr){2-4} \cmidrule(lr){5-7} \cmidrule(lr){8-10}
   & {ACC $\uparrow$} & {TOK $\downarrow$} &{EFF $\uparrow$} & {ACC $\uparrow$} & {TOK $\downarrow$} &{EFF $\uparrow$}& {ACC $\uparrow$} & {TOK $\downarrow$} &{EFF $\uparrow$} \\ 
\midrule
\makecell{Qwen2.5-32B-\\Instruct} & 15.8 & 954 & 1.66E-04 & 67.2 & 737 & 9.11E-04 & 47.0 & 517 & 9.08E-04 \\
\midrule
\makecell{R1-Distill-\\Qwen-32B} & 69.2 & 9,311 & 7.43E-05 & 94.4 & 5,561 & 1.70E-04 & 64.7 & 5,634 & 1.15E-04 \\
\midrule
QwQ & 81.7 & 12,234 & 6.68E-05 & 97.8 & 7,350 & 1.33E-04 & 70.2 & 7,483 & 9.38E-05 \\
\midrule
\midrule
S1-32B & 37.9 & 6,646 & 5.71E-05 & 80.9 & 4,542 & 1.78E-04 & 60.7 & 4,172 & 1.46E-04 \\
S1-32B-P & \textbf{\annotate{42.1}{red}{+4.2}} & \textbf{\annotate{4,716}{blue}{-29\%}} & \textbf{\annotate{8.92E-05}{red}{+56\%}} & \textbf{\annotate{83.1}{red}{+2.2}} & \textbf{\annotate{3,809}{blue}{-16\%}} & \textbf{\annotate{2.18E-04}{red}{+22\%}} & \textbf{\annotate{61.6}{red}{+0.9}} & \textbf{\annotate{2,472}{blue}{-41\%}} & \textbf{\annotate{2.49E-04}{red}{+71\%}} \\
\midrule
\midrule
LIMO & 56.7 & 12,497 & 4.53E-05 & 91.9 & 5,516 & 1.67E-04 & 67.2 & 7,173 & 9.36E-05 \\
LIMO-P & \textbf{\annotate{63.3}{red}{+6.6}} & \textbf{\annotate{10,588}{blue}{-15\%}} & \textbf{\annotate{5.98E-05}{red}{+32\%}} & \textbf{\annotate{93.8}{red}{+1.9}} & \textbf{\annotate{5,235}{blue}{-5\%}} & \textbf{\annotate{1.79E-04}{red}{+7\%}} & \textbf{\annotate{71.2}{red}{+4}} & \textbf{\annotate{6,969}{blue}{-3\%}} & \textbf{\annotate{1.02E-04}{red}{+9\%}} \\
\midrule
\midrule
LIMO-V2 & 66.3 & 13,896 & 4.77E-05 & 94.4 & 6,843 & 1.38E-04 & 70.2 & 8,035 & 8.74E-05 \\
LIMO-V2-P & \textbf{\annotate{71.2}{red}{+4.9}} & \textbf{\annotate{12,163}{blue}{-12\%}} & \textbf{\annotate{5.86E-05}{red}{+23\%}} & \textbf{\annotate{96.6}{red}{+2.2}} & \textbf{\annotate{6,348}{blue}{-7\%}} & \textbf{\annotate{1.52E-04}{red}{+10\%}} & \textbf{\annotate{74.2}{red}{+3}} & \textbf{\annotate{6,968}{blue}{-13\%}} & \textbf{\annotate{1.07E-04}{red}{+22\%}} \\
\bottomrule
\end{tabular}
}
\vspace{-10pt}
\end{table}

\subsection{Main Results}

\paragraph{Performance Across Benchmarks} 
Table~\ref{main-exp} demonstrates the consistent effectiveness of our PIR optimization framework across multiple challenging reasoning benchmarks. The PIR-optimized variants achieve superior efficiency-accuracy trade-offs across all model families. S1-32B-P shows remarkable improvements on AIME with a 4.2 percentage point accuracy increase alongside a 29\% token reduction, yielding a 56\% efficiency improvement. Similarly, LIMO-P demonstrates consistent gains with enhanced accuracy (+6.6, +1.9, and +4.0 percentage points across AIME, AMC, and GPQA Diamond, respectively) while reducing token consumption by up to 15\%, achieving efficiency improvements of 32\%, 7\%, and 9\% across the three benchmarks. LIMO-V2-P exhibits comparable enhancements with token reductions of 12\%, 7\%, and 13\%, paired with accuracy improvements of 4.9, 2.2, and 3.0 percentage points. These consistent improvements across diverse benchmarks confirm that our PIR framework effectively identifies and preserves high-value reasoning steps while eliminating low-importance functional components.

\paragraph{Generalizability to Different Data Sources} The substantial performance improvements across models trained on data distilled from diverse foundation models (S1 from Gemini Flash Thinking, LIMO from DeepSeek-R1, and LIMO-V2 from QwQ) suggest PIR's strong generalizability potential. While individual gains vary in magnitude, the consistent pattern of simultaneous accuracy increases and token reductions across these different data sources indicates that our framework successfully identifies important reasoning patterns independent of the original distillation source. This suggests that PIR captures fundamental aspects of reasoning quality rather than exploiting source-specific characteristics. 
\paragraph{Comparison with Alternative Optimization Methods}
Table~\ref{tab:s1-ablation} demonstrates our PIR framework's superior performance compared to other reasoning optimization approaches. PIR (S1-32B-P) consistently outperforms all alternatives across benchmarks, achieving significant improvements over the baseline S1-32B model with accuracy gains of 4.2, 2.2, and 0.9 percentage points alongside token reductions of 29\%, 16\%, and 41\% on AIME, AMC, and GPQA Diamond respectively. Notably, our approach of selectively pruning only functional steps while preserving all progressive reasoning components proves superior to whole-sentence pruning methods (S1-SPIRIT), which apply filtering indiscriminately across all reasoning steps. The superior accuracy of S1-32B-P over S1-SPIRIT, particularly on AIME (5.0 percentage point advantage), empirically validates our hypothesis that progressive reasoning steps constitute the essential solution backbone and should remain intact.

\begin{table}[t]
\vspace{-15pt}  
\centering
\setlength{\tabcolsep}{1.8pt}
\renewcommand{\arraystretch}{1.25}
\caption{Experimental results comparing PIR(S1-32B-P) with different optimization approaches.}
\resizebox{\linewidth}{!}{%
\begin{tabular}{lllllllllll}
\toprule
\multirow{2}{*}{\textbf{\makecell{Model/\\Method}}}& \multirow{2}{*}{\textbf{\makecell{Training\\Tokens}}}  &\multicolumn{3}{c}{\textbf{AIME}} & \multicolumn{3}{c}{\textbf{AMC}} & \multicolumn{3}{c}{\textbf{GPQA Diamond}} \\
\cmidrule(lr){3-5} \cmidrule(lr){6-8} \cmidrule(lr){9-11}
&&{ACC $\uparrow$} & {TOK $\downarrow$} & {EFF $\uparrow$} & {ACC $\uparrow$} & {TOK $\downarrow$} & {EFF $\uparrow$} & {ACC $\uparrow$} & {TOK $\downarrow$} & {EFF $\uparrow$} \\
\midrule
S1-32B &4.51E+06 &37.9 & 6,646 & 5.71E-05 & 80.9 & 4,542 & 1.78E-04&60.7& 4,172&1.46E-04 \\
S1-PROMPT&4.51E+06 & 36.7 & 8,013 & 4.58E-05 & 72.5 & 3,724 & 1.95E-04 &58.0&2,853&2.03E-04\\
S1-SPIRIT&4.32E+06 & 37.1 & 4,906 & 7.56E-05 & 81.3 & 3,517 & 2.31E-04 &60.1&2,818&2.13E-04\\
S1-RULE&4.31E+06 & 36.7 & 4,807 & 7.63E-05 & 81.3 & 3,654 & 2.22E-04&
58.1&3,837&1.51E-04\\
S1-32B-P&4.31E+06  & \textbf{\annotate{42.1}{red}{+4.2}} & \textbf{\annotate{4,716}{blue}{-29\%}} & \textbf{\annotate{8.92E-05}{red}{+56\%}} & \textbf{\annotate{83.1}{red}{+2.2}} & \textbf{\annotate{3,809}{blue}{-16\%}} & \textbf{\annotate{2.18E-04}{red}{+22\%}}&
\textbf{\annotate{61.6}{red}{+0.9}} & \textbf{\annotate{2,472}{blue}{-41\%}} & \textbf{\annotate{2.49E-04}{red}{+71\%}}
\\
\bottomrule
\end{tabular}
}
\label{tab:s1-ablation}
\vspace{-20pt}  
\end{table}




\subsection{Analysis}

\subsubsection{Generalizability to Test-Time Scaling}

Test-time scaling represents a critical dimension of LLM reasoning deployment, where more computational resources are allocated during inference. To evaluate whether our PIR optimization framework maintains its effectiveness across varying inference-time token budgets, we conducted experiments measuring accuracy as a function of token budget constraints.
Figure~\ref{fig:cover} presents the test-time scaling curves across three benchmarks of LIMO and LIMO-P. The results demonstrate that PIR-optimized models consistently outperform their non-optimized counterparts across most token budget levels. The superior performance of PIR-optimized models across different token budgets demonstrates the generalizability of our approach to different resource constraints. This makes our PIR framework particularly valuable for real-world applications where inference efficiency and response latency are crucial considerations. The ability to maintain performance advantages highlights the robustness of our optimization methodology to different deployment scenarios.

\subsubsection{Impact of ratios}

\begin{figure}[t]
    \centering
    \includegraphics[width=0.9\linewidth]{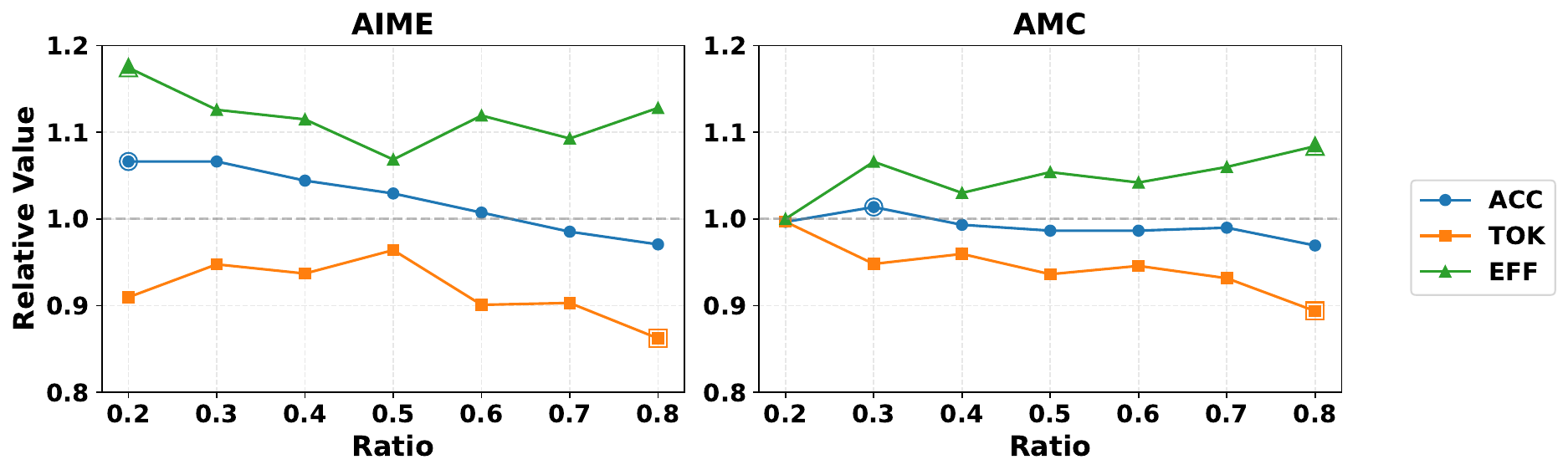}
    \caption{Impact of pruning ratio on model performance. This figure displays relative performance metrics (normalized to baseline) across different pruning ratios for AIME and AMC benchmarks. The horizontal dashed line represents the baseline performance (ratio=0). }
    \label{fig:ration}
    \vspace{-20pt}
\end{figure}

We investigated the impact of various pruning ratios on model performance by creating multiple LIMO dataset variants with different proportions of functional reasoning steps removed. As shown in Figure \ref{fig:ration}, our experiments revealed clear performance trade-offs across metrics. For the AIME benchmark, lower pruning ratios (0.2-0.3) yielded optimal accuracy improvements over the baseline, while higher ratios (0.8) achieved the greatest response length reduction. For AMC, a moderate pruning ratio of 0.3 delivered peak accuracy while maintaining efficiency gains. Test time efficiency consistently improved with pruning across both benchmarks, with particularly strong gains on AIME. These results demonstrate the existence of an optimal refinement threshold that balances the removal of redundant functional steps with the preservation of critical reasoning components. Excessive pruning beyond this threshold leads to declining accuracy as valuable reasoning elements are removed, even when categorized as functional steps. This pattern validates our PIR framework's approach of selectively identifying and preserving high-value reasoning steps based on their quantitative importance scores.

\subsubsection{Generalizability to Model Size}
\begin{figure}
    \centering
    \includegraphics[width=\linewidth]{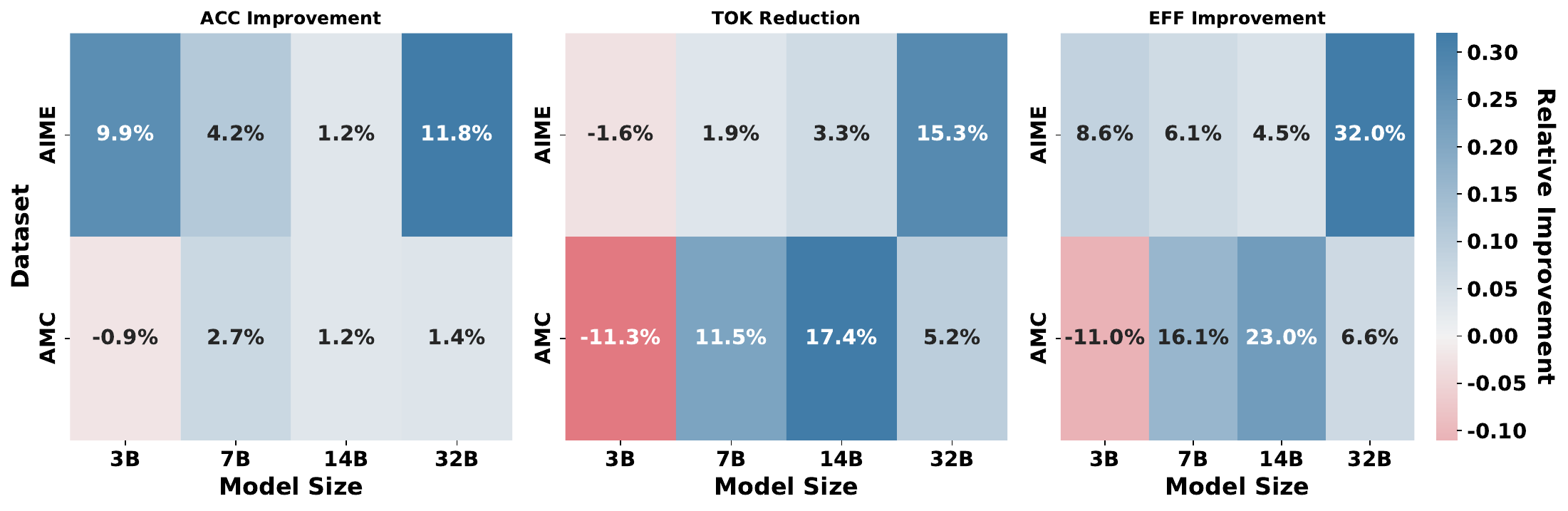}
    \caption{Impact of PIR refinement across model sizes and benchmarks. Heatmaps show relative percentage changes between models trained with pruned versus original datasets. Blue indicates improvement: higher accuracy, shorter response length, or better efficiency. }
    \label{fig:model size}
    \vspace{-15pt}
\end{figure}

To evaluate PIR's generalizability across parameter scales, we conducted experiments using Qwen2.5 models ranging from 3B to 32B parameters, comparing the performance of models trained with optimized LIMO-P versus original LIMO datasets. As shown in Figure \ref{fig:model size}, our method demonstrates robust scalability with performance improvements across most model sizes. The benefits of PIR refinement become increasingly pronounced as model size increases, particularly for the AIME benchmark where the 32B model shows impressive gains across all metrics (11.8\% accuracy improvement, 15.3\% response length reduction, and 32.0\% efficiency increase). For AMC, mid-sized models (7B-14B) yield the strongest efficiency improvements (up to 23.0\%). The consistent pattern of enhancement for most model sizes suggests that our method's scalability and practical utility across various model sizes.




%% file: chapter/related_work.tex
\section{Related Work}

\subsection{Test-Time Scaling of LLMs}

Test-time scaling \citep{jaech2024openai, snell2024scaling, qwq32b, guo2025deepseek} enhances LLM reasoning by increasing inference-time computation. Approaches include non-training methods, which optimize existing model inference strategies, and training-based methods, which modify model parameters. Non-training techniques encompass Best-of-N sampling \citep{lightman2023let,song2024good}, majority voting \citep{wang2022self,chen2024more}, and tree search \citep{yao2023tree,tian2024toward} for exploring multiple reasoning paths. Training-based approaches divide into supervised fine-tuning (SFT) and reinforcement learning (RL). SFT methods train on high-quality reasoning traces \citep{muennighoff2025s1,ye2025limo,li2025llms,bespoke_stratos}, with S1 and LIMO demonstrating improved performance through careful sample selection. RL approaches \citep{schulman2017proximal, guo2025deepseek,jaech2024openai,qwq32b} have yielded exceptional results, with DeepSeek-R1 using GRPO and OpenAI's o1 and QwQ enabling autonomous development of reasoning chains that adaptively allocate computation based on problem complexity. Our work focuses on optimizing SFT-based approaches through perplexity-based refinement to improve efficiency while preserving accuracy.

\subsection{Efficient Reasoning}

Research on efficient reasoning has gained significant traction as LLMs face challenges with computational overhead and verbosity. At inference time, various optimization approaches have emerged without requiring parameter updates. Length budgeting techniques like Token-Budget-Aware LLM Reasoning \citep{han2024token} enforce token limits via prompting, while S1 \citep{muennighoff2025s1} appends end-of-thinking delimiters. Dual-process inspired system switching methods alternate between fast intuitive and deliberative reasoning; Dualformer \citep{su2024dualformer} selectively drops reasoning traces during training, while System 1.x \citep{saha2024system} employs a controller to assess task difficulty. Model switching approaches such as BiLD \citep{kim2023speculative} and EAGLE \citep{li2024eagle} leverage speculative decoding with smaller models for initial predictions, while RouteLLM \citep{ong2406routellm} routes queries based on complexity. For supervised fine-tuning, C3ot \citep{kang2025c3ot} preserves essential information while reducing redundancies, and TokenSkip \citep{xia2025tokenskip} omits less important tokens. SPIRIT \citep{cui2025stepwise} prunes low-importance reasoning steps. RL-based methods either incorporate explicit length penalties \citep{luo2025o1,shen2025dast} or balance exploitation with exploration \citep{qu2025optimizing}. Most closely related to our work, SPIRIT \citep{cui2025stepwise} calculates perplexity for all reasoning steps and filters them based on a predetermined ratio. Our approach fundamentally differs by distinguishing between reasoning step types rather than treating all equally. We preserve all progressive reasoning steps—the essential solution backbone—while only pruning less critical functional components (verification, multi-method validation, and error correction), ensuring core reasoning integrity and avoiding the risk of removing critical solution elements.

\section{Conclusion}
\paragraph{Contributions}This paper introduces PIR (Perplexity-based Importance Refinement), a novel framework that optimizes reasoning chains by quantitatively assessing step importance and selectively pruning low-value functional elements while preserving essential reasoning paths. Our comprehensive evaluation demonstrates that models fine-tuned on PIR-optimized datasets achieve both improved accuracy and significantly reduced token usage. By strategically balancing thorough problem-solving with computational efficiency, PIR establishes a principled approach for deploying advanced reasoning capabilities in latency-sensitive applications, opening new avenues for research on efficient reasoning in foundation models.

\paragraph{Limitations}While our approach demonstrates significant improvements, several limitations warrant further investigation. First, our evaluation primarily focuses on mathematical reasoning tasks and science tasks; future work should validate PIR's effectiveness across broader reasoning domains including logical, commonsense, and causal reasoning. Second, our refinement strategy relies on perplexity as the primary importance indicator, which may not fully capture the semantic contribution of certain reasoning steps. Alternative metrics incorporating semantic relevance could enhance refinement precision. Third, the optimal pruning ratio may vary across different reasoning tasks and model architectures, suggesting the need for adaptive refinement strategies. Finally, our approach currently requires access to model perplexity outputs, which may limit applicability with closed-source models. Addressing these limitations could further advance efficient reasoning frameworks for real-world applications.

%% file: chapter/acknowledgement.tex
\section{Acknowledgements}
We would like to thank all reviewers for their insightful comments and suggestions to help improve the paper. This work was partially supported by the Research Grants Council of Hong Kong (GRF No. 15209724). This work was also partially funded by the National Natural Science Foundation of China (62476168) and SII.

%% file: chapter/append.tex
\section{Reasoning Refinement: Reasoning Optimization Framework}
\label{append:pattern case}

\subsection{Phrases Used for Pattern Matching}

\begin{figure}[ht]
    \centering
    \includegraphics[width=1\linewidth]{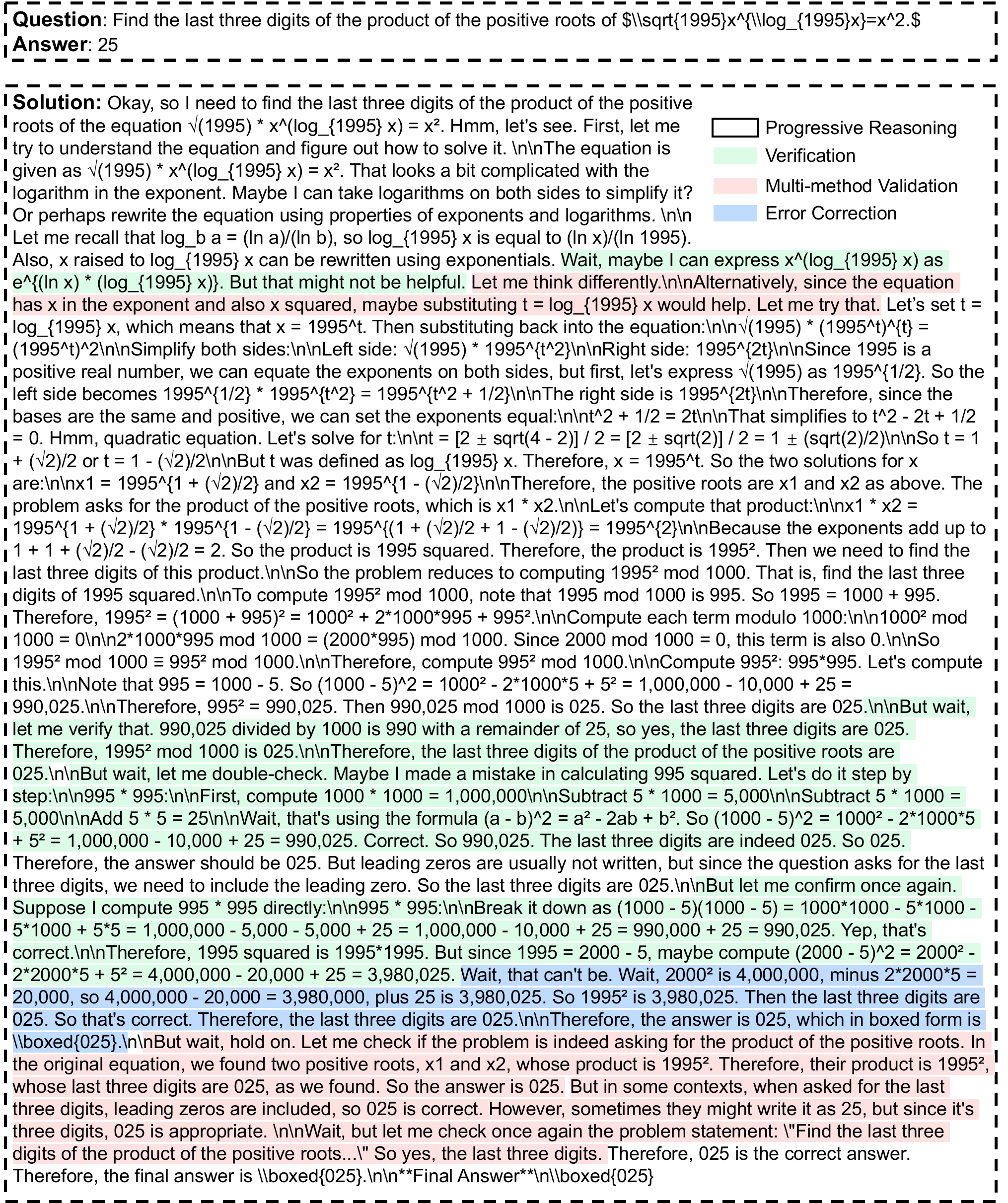}
    \caption{A case where one training sample contains the four patterns.}
    \label{fig:pattern_case}
\end{figure}

In our rule-based pattern matching system for reasoning step classification, we identify four distinct cognitive reasoning patterns characterized by specific linguistic markers. Table \ref{tab:reasoning_patterns} presents these patterns alongside their characteristic phrases. Figure \ref{fig:pattern_case} demonstrates one sample from S1K containing the four patterns.

\textbf{Progressive Reasoning} constitutes the critical path to solution derivation, characterized by forward-chaining inference that follows a deductive logical progression from premises to conclusion. This pattern forms the essential backbone of solution development and is indicated by phrases that signal logical advancement, such as ``Let's solve,'' ``First/Then/Next,'' and ``Therefore.''

\textbf{Verification} represents metacognitive monitoring processes where previous calculations are systematically validated for accuracy. This pattern is identified through phrases like ``Let me check,'' ``Let me verify,'' and ``Double-check,'' which signal when the model reviews its prior work.

\textbf{Multi-method Validation} demonstrates convergent thinking through the application of diverse methodological approaches to reinforce conclusions. This pattern is recognized through expressions indicating alternative approach consideration, including ``Alternatively,'' ``Another way,'' and ``Let's try a different approach,'' potentially introducing redundancy.

\textbf{Error Correction} embodies a self-regulatory mechanism through which logical inconsistencies or computational errors are identified and remediated. This pattern is captured by phrases acknowledging mistakes, such as ``This is wrong,'' ``The mistake was,'' and ``This contradicts.''

\begin{table}[ht]
\centering
\caption{Cognitive Reasoning Patterns and Associated Phrases}
\begin{tabular}{p{0.45\textwidth}p{0.45\textwidth}}
\toprule
\textbf{Cognitive Reasoning Patterns} & \textbf{Phrases} \\
\midrule
\begin{tabular}[c]{@{}p{0.45\textwidth}@{}} Progressive Reasoning \end{tabular} & 
\begin{tabular}[c]{@{}p{0.45\textwidth}@{}}
``Let's solve''\\
``First/Then/Next''\\
``Therefore''\\
``We need to''\\
``Given that''
\end{tabular} \\
\midrule
\begin{tabular}[c]{@{}p{0.45\textwidth}@{}} Verification \end{tabular} & 
\begin{tabular}[c]{@{}p{0.45\textwidth}@{}}
``Wait''\\
``Let me check''\\
``Let me verify''\\
``Double-check''\\
``Going back to''
\end{tabular} \\
\midrule
\begin{tabular}[c]{@{}p{0.45\textwidth}@{}} Multi-method Validation \end{tabular} & 
\begin{tabular}[c]{@{}p{0.45\textwidth}@{}}
``Alternatively''\\
``Another way''\\
``Let's try a different approach''\\
``Using another method''\\
``We can also verify''
\end{tabular} \\
\midrule
\begin{tabular}[c]{@{}p{0.45\textwidth}@{}} Error Correction \end{tabular} & 
\begin{tabular}[c]{@{}p{0.45\textwidth}@{}}
``This is wrong''\\
``The mistake was''\\
``That's impossible''\\
``This contradicts''\\
``The error is''
\end{tabular} \\
\bottomrule
\end{tabular}

\label{tab:reasoning_patterns}
\end{table}

\subsection{Prompt used for Pattern Recognition}

\begin{figure}[ht]
    \centering
    \includegraphics[width=1\linewidth]{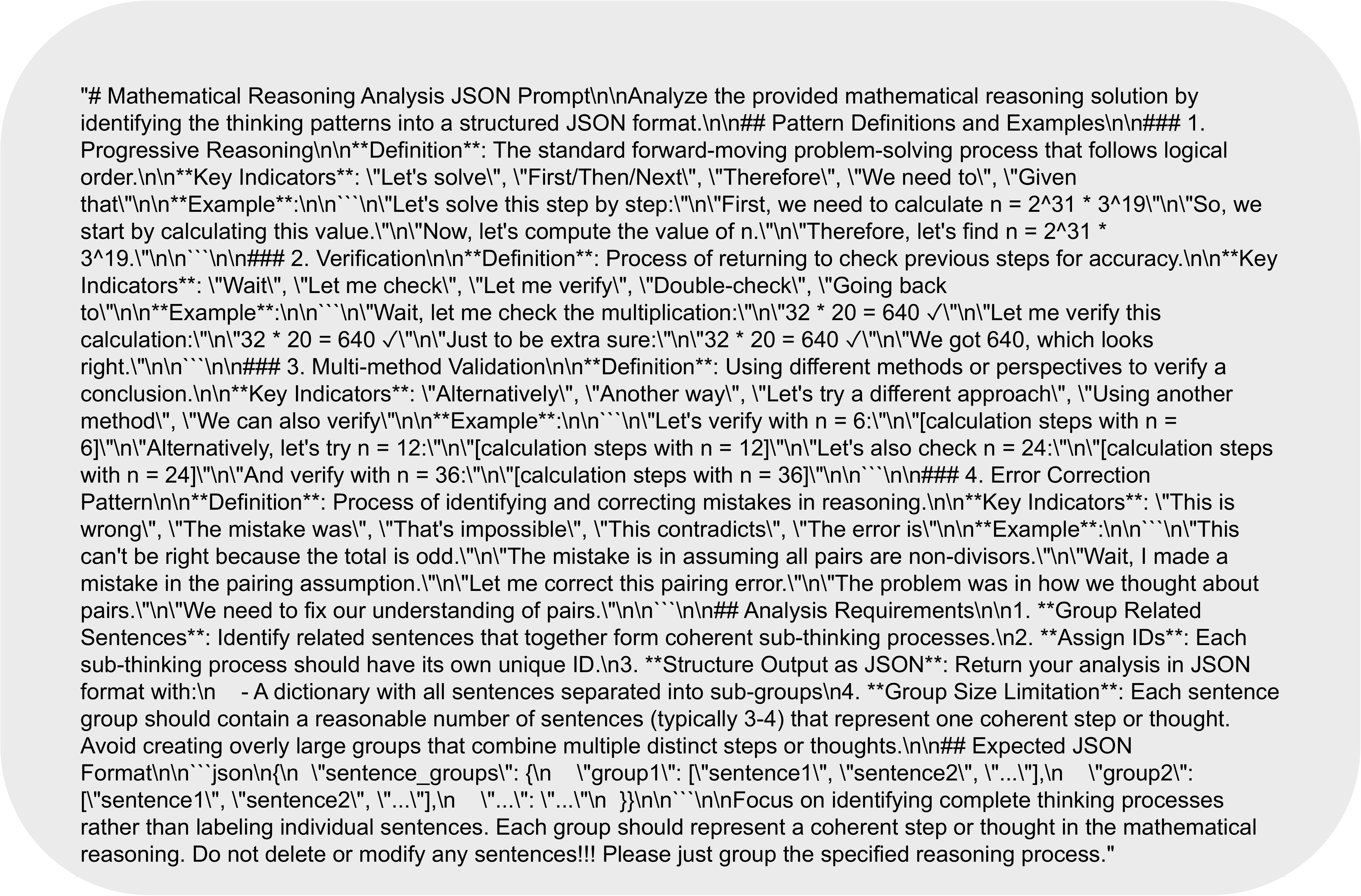}
    \caption{The prompt to segment coherent sub-thinking sentences into cohesive reasoning steps.}
    \label{fig:prompt_catogory}
\end{figure}

\begin{figure}[ht]
    \centering
    \includegraphics[width=1\linewidth]{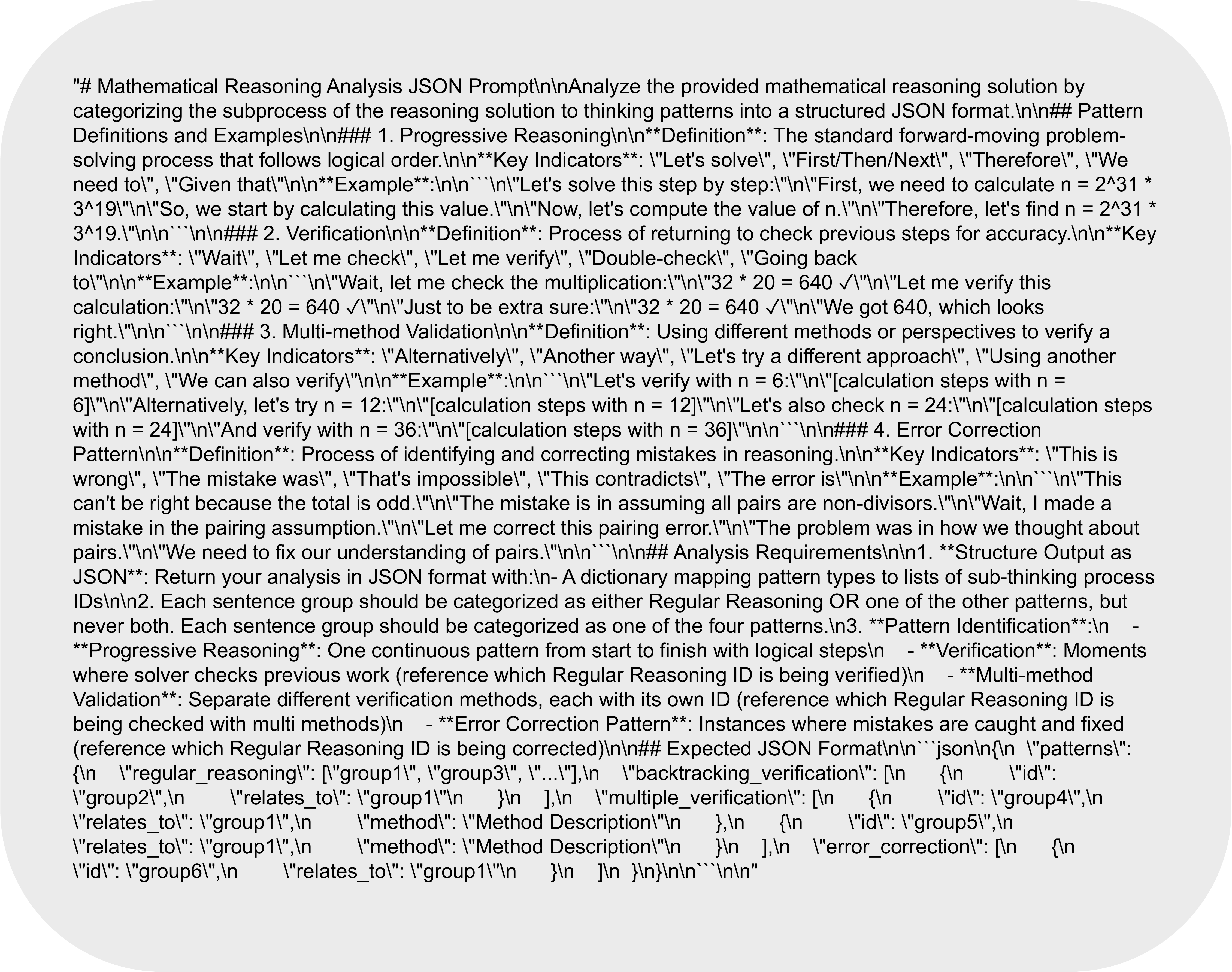}
    \caption{The prompt to categorize the steps into reasoning patterns}
    \label{fig:prompt_class}
\end{figure}

Our hybrid classification system combines rule-based pattern matching with contextual analysis using Claude 3.7 Sonnet for more nuanced reasoning pattern recognition. In Figure \ref{fig:prompt_catogory} and \ref{fig:prompt_class}, we present the prompts used for the sophisticated pattern recognition of Claude 3.7 Sonnet.

\subsection{The Pseudocode of the Pipeline of PIR}

Algorithm \ref{alg:alg1} outlines the complete pipeline of our Perplexity-based Importance Refinement (PIR) framework. Our PIR framework implements a streamlined three-phase process for optimizing reasoning chains. First, Claude 3.7 Sonnet performs hierarchical decomposition of reasoning into coherent steps, which are classified using our hybrid system combining rule-based pattern matching and contextual analysis. Second, we quantify each functional step's importance by calculating its PIR value—the logarithmic ratio between the perplexity of generating the answer with and without that step using Qwen2.5-32B-Instruct. Finally, we apply pattern-specific selective pruning, preserving all progressive reasoning while removing only low-value functional components (verification, multi-method validation, and error correction) based on their PIR scores, thereby maintaining solution integrity while significantly reducing verbosity.

\begin{algorithm}[ht]
\caption{Reasoning Chain Optimization via PIR}
\label{alg:alg1}
\resizebox{0.8\textwidth}{!}{
\begin{minipage}{\textwidth}
\SetAlgoLined
\KwIn{$solution$: original reasoning chain, $answer$: solution answer, $ratio$: pruning threshold, $\theta$: evaluation model}
\KwOut{$solution_{opt}$: optimized reasoning chain}

\vspace{1mm}
\texttt{// Step 1: Segment and classify reasoning chain steps}\\
$steps \leftarrow \text{SegmentIntoLogicalSteps}(solution, \theta)$\; 
$classified\_steps \leftarrow \text{ClassifyReasoningPatterns}(steps, \theta)$\; 
$functional\_steps \leftarrow \text{FilterByPatterns}(classified\_steps, \{\text{Verification, Multi-method, Error correction}\})$\; 

\vspace{1mm}
\texttt{// Step 2: Calculate baseline perplexity with complete reasoning}\\
$PPL_{\theta}(R) \leftarrow \text{CalculatePerplexity}(solution, answer, \theta)$\;

\vspace{1mm}
\texttt{// Step 3: Evaluate importance of each functional step}\\
\ForEach{$step_i \in functional\_steps$}{
    $solution_{-i} \leftarrow \text{RemoveStep}(solution, step_i)$\;
    $PPL_{\theta}(R \setminus \{step_i\}) \leftarrow \text{CalculatePerplexity}(solution_{-i}, answer, \theta)$\;
    $step_i.PIR \leftarrow \log\left(\frac{PPL_{\theta}(R \setminus \{step_i\})}{PPL_{\theta}(R)}\right)$\;
}

\vspace{1mm}
\texttt{// Step 4: Selectively prune low-importance steps by pattern type}\\
$solution_{opt} \leftarrow solution$\;
\ForEach{$pattern \in \{\text{Verification, Multi-method, Error correction}\}$}{
    $pattern\_steps \leftarrow \text{FilterByPattern}(functional\_steps, pattern)$\;
    $threshold \leftarrow \text{CalculatePruningThreshold}(pattern\_steps, ratio)$\;
    $steps\_to\_prune \leftarrow \text{SelectLowPIRSteps}(pattern\_steps, threshold)$\;
    $solution_{opt} \leftarrow \text{RemoveSteps}(solution_{opt}, steps\_to\_prune)$\;
}

\Return{$solution_{opt}$}\;
\end{minipage}
}
\end{algorithm}
\vspace{-10pt}

\section{Experiments}

\subsection{Dataset Statistics}
\label{append:full_data_statistic}
This section provides comprehensive statistics for the datasets used in our experiments before and after applying the PIR optimization framework at different thresholds. Table~\ref{table:full_data_static} presents the number of examples and total token counts for each dataset variant.
The original datasets (S1, LIMO, and LIMO-V2) were derived from three different Large Reasoning Models: Gemini Flash Thinking, DeepSeek-R1, and QwQ, respectively. Each dataset was then processed using our PIR framework at varying optimization ratios (from 0.2 to 0.8), where higher values indicate more aggressive pruning of functional reasoning steps.

\begin{table}[ht]
\centering
\caption{Statistics of Dataset Variants after PIR Optimization}
\label{table:full_data_static}
\begin{tabular}{c|c|c|c}
\toprule
\textbf{Source} & \textbf{Data} & \textbf{Numbers} & \textbf{Tokens} \\
\midrule
\multirow{8}{*}{Gemini} & S1 & 1000 & 4509505 \\
 & S1-0.2 & 1000 & 4440447 \\
 & S1-0.3 & 1000 & 4390349 \\
 & S1-0.4 & 1000 & 4307878 \\
 & S1-0.5 & 1000 & 4173726 \\
 & S1-0.6 & 1000 & 4136261 \\
 & S1-0.7 & 1000 & 4064908 \\
 & S1-0.8 & 1000 & 3998891 \\
\midrule
\multirow{8}{*}{DeepSeek-R1} & LIMO & 817 & 5144004 \\
 & LIMO-0.2 & 817 & 4971633 \\
 & LIMO-0.3 & 817 & 4865402 \\
 & LIMO-0.4 & 817 & 4724104 \\
 & LIMO-0.5 & 817 & 4542101 \\
 & LIMO-0.6 & 817 & 4459545 \\
 & LIMO-0.7 & 817 & 4342583 \\
 & LIMO-0.8 & 817 & 4220217 \\
\midrule
\multirow{8}{*}{QwQ} & LIMO-V2 & 800 & 8866950 \\
 & LIMO-V2-0.2 & 800 & 8488260 \\
 & LIMO-V2-0.3 & 800 & 8398975 \\
 & LIMO-V2-0.4 & 800 & 8292603 \\
 & LIMO-V2-0.5 & 800 & 8161582 \\
 & LIMO-V2-0.6 & 800 & 8082063 \\
 & LIMO-V2-0.7 & 800 & 7980495 \\
 & LIMO-V2-0.8 & 800 & 7877110 \\
\bottomrule

\end{tabular}
\end{table}

\subsection{Main Results with Training Tokens}
Table~\ref{tab:model_comparison} presents a comprehensive evaluation of our PIR optimization approach across three challenging benchmarks: AIME (American Invitational Mathematics Examination), AMC (American Mathematics Competition), and GPQA Diamond. We include training token counts alongside accuracy, response length, and test-time efficiency to highlight the relationship between data efficiency and model performance.
The baseline models—R1-Distill-Qwen-32B, Qwen2.5-32B-Instruct, and QWQ—establish performance references across the benchmarks. QWQ demonstrates superior accuracy but requires substantially longer responses, which impacts its test-time efficiency. Qwen2.5-32B-Instruct offers the highest efficiency due to its concise responses, albeit with lower accuracy.
For our optimized models, we observe consistent patterns across all three dataset families (S1, LIMO, and LIMO-V2): \textbf{S1 datasets:} The PIR-optimized variant (S1-32B-P) achieves higher accuracy on AIME (+4.2\%) and AMC (+2.2\%) while using 4.4\% fewer training tokens. The average response length decreases by 29\% for AIME and 16\% for AMC, resulting in efficiency improvements of 56\% and 22\%, respectively. \textbf{LIMO datasets:} The PIR-optimized model (LIMO-P) demonstrates accuracy improvements across all benchmarks (+6.6\% on AIME, +1.9\% on AMC, +4\% on GPQA) while requiring 8.8\% fewer training tokens. Response length reductions of 15\% for AIME and 5\% for AMC translate to efficiency gains of 32\% and 7\%, respectively. 
\textbf{LIMO-V2 datasets:} PIR optimization (LIMO-V2-P) achieves the consistent accuracy improvements (+4.9\% on AIME, +2.2\% on AMC, +3\% on GPQA) while using 5.3\% fewer training tokens. Response lengths decrease consistently (12\% for AIME, 7\% for AMC, 13\% for GPQA), yielding efficiency improvements of 23\%, 10\%, and 22\%, respectively.
These results demonstrate that our PIR framework effectively reduces training token requirements while simultaneously improving both accuracy and efficiency across diverse reasoning tasks. The consistent performance improvements across different model families validate the generalizability of our approach. Notably, the LIMO-V2-P model achieves state-of-the-art performance on all benchmarks while maintaining competitive efficiency, highlighting the effectiveness of optimizing reasoning chains by preserving essential progressive reasoning while removing less valuable functional steps.

\begin{table*}[ht]
\centering
\caption{Performance comparison across different models on AIME, AMC, and GPQA Diamond benchmarks.}
\label{tab:model_comparison}
\resizebox{\textwidth}{!}{%
\begin{tabular}{ccccccccccccc}
\toprule
\multirow{2}{*}{Model} & \multicolumn{4}{c}{AIME} & \multicolumn{4}{c}{AMC} & \multicolumn{4}{c}{GPQA Diamond} \\
\cmidrule(lr){2-5} \cmidrule(lr){6-9} \cmidrule(lr){10-13}
 & \makecell{Training\\Tokens} & Acc & \makecell{Avg. Response\\Tokens} & \makecell{Test Time\\Efficiency} & \makecell{Training\\Tokens} & Acc & \makecell{Avg. Response\\Tokens} & \makecell{Test Time\\Efficiency} & \makecell{Training\\Tokens} & Acc & \makecell{Avg. Response\\Tokens} & \makecell{Test Time\\Efficiency} \\
\midrule
\makecell{Qwen2.5-32B-\\Instruct} & N/A & 15.8 & 954 & 1.66E-04 & N/A & 67.2 & 737 & 9.11E-04 & N/A & 47.0 & 517 & 9.08E-04 \\
\makecell{R1-Distill-\\Qwen-32B} &N/A & 69.2 & 9,311 & 7.43E-05 & N/A & 94.4 & 5,561 & 1.70E-04 & N/A & 64.7 & 5,634 & 1.15E-04 \\
QWQ & N/A & 81.7 & 12,234 & 6.68E-05 & N/A & 97.8 & 7,350 & 1.33E-04 & N/A & 70.2 & 7,483 & 9.38E-05 \\
\midrule
\multicolumn{13}{c}{S1} \\
\midrule
S1-32B & 4.51E+06 & 37.9 & 6,646 & 5.71E-05 & 4.51E+06 & 80.9 & 4,542 & 1.78E-04 & 4.51E+06 & 60.7 & 4,172 & 1.46E-04 \\
S1-32B-P & 4.31E+06 & 42.1 & 4,716 & 8.92E-05 & 4.31E+06 & 83.1 & 3,809 & 2.18E-04 & 4.39E+06 & 61.6 & 2,472 & 2.49E-04 \\
\midrule
\multicolumn{13}{c}{LIMO} \\
\midrule
LIMO & 5.14E+06 & 56.7 & 12,497 & 4.53E-05 & 5.14E+06 & 91.9 & 5,516 & 1.67E-04 & 5.14E+06 & 67.2 & 7,173 & 9.36E-05 \\
LIMO-P & 4.69E+06 & 63.3 & 10,588 & 5.98E-05 & 4.78E+06 & 93.8 & 5,235 & 1.79E-04 & 4.72E+06 & 71.2 & 6,969 & 1.02E-04 \\
\midrule
\multicolumn{13}{c}{LIMO-V2} \\
\midrule
LIMO-V2 & 8.87E+06 & 66.3 & 13,896 & 4.77E-05 & 8.87E+06 & 94.4 & 6,843 & 1.38E-04 & 8.87E+06 & 70.2 & 8,035 & 8.74E-05 \\
LIMO-V2-P & 8.40E+06 & 71.2 & 12,163 & 5.65E-05 & 8.40E+06 & 96.6 & 6,348 & 1.52E-04 & 8.49E+06 & 74.2 & 6,968 & 1.07E-04 \\
\bottomrule
\end{tabular}%
}
\end{table*}

\subsection{Impact of Cognitive Reasoning Patterns}

\begin{figure} 
\centering
\setlength{\tabcolsep}{1.8pt}
\includegraphics[width=\linewidth]{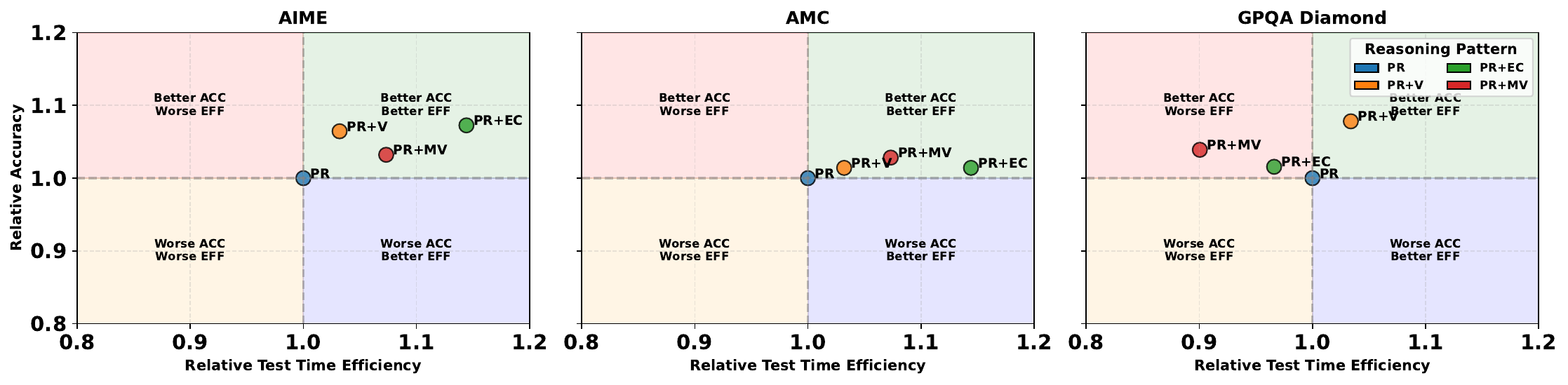}
\caption{Impact of reasoning patterns on model performance across different benchmarks. Each subplot displays the relative accuracy (y-axis) versus relative test time efficiency (x-axis) compared to the Progressive Reasoning (PR) baseline. PR represents the model trained with only progressive reasoning steps. PR+Verification (PR+V) is trained with the dataset that includes progressive reasoning and verification steps. PR+Error Correction (PR+EC) stands for the model trained with progressive reasoning and error correction steps. PR+Multi-method Validation (PR+MV) is trained with progressive reasoning and multi-method validation steps.}
\vspace{-10pt}  
\label{fig:type}
\end{figure}

Our empirical analysis of cognitive reasoning patterns reveals distinct performance characteristics across the evaluated benchmarks. As shown in Figure \ref{fig:type}, the four identified reasoning patterns exhibit different trade-offs between accuracy and computational efficiency. Progressive Reasoning (PR) provides a solid baseline, while PR+Error Correction demonstrates the most balanced performance improvement on several datasets, delivering notable accuracy gains with competitive or even improved efficiency. These findings validate our PIR framework's approach of quantitatively evaluating reasoning step importance to identify and selectively preserve high-value steps while pruning those with minimal contribution, creating optimized reasoning chains that balance accuracy gains with computational demands. The observed performance variations highlight that different reasoning patterns contribute differentially to model performance, with certain patterns delivering more substantial benefits in specific contexts. This suggests that selectively preserving the most valuable functional reasoning components while removing redundant steps can effectively optimize the efficiency-accuracy trade-off in reasoning chains.

\subsection{Case Study}
\begin{figure}
    \centering
    \includegraphics[width=1\linewidth]{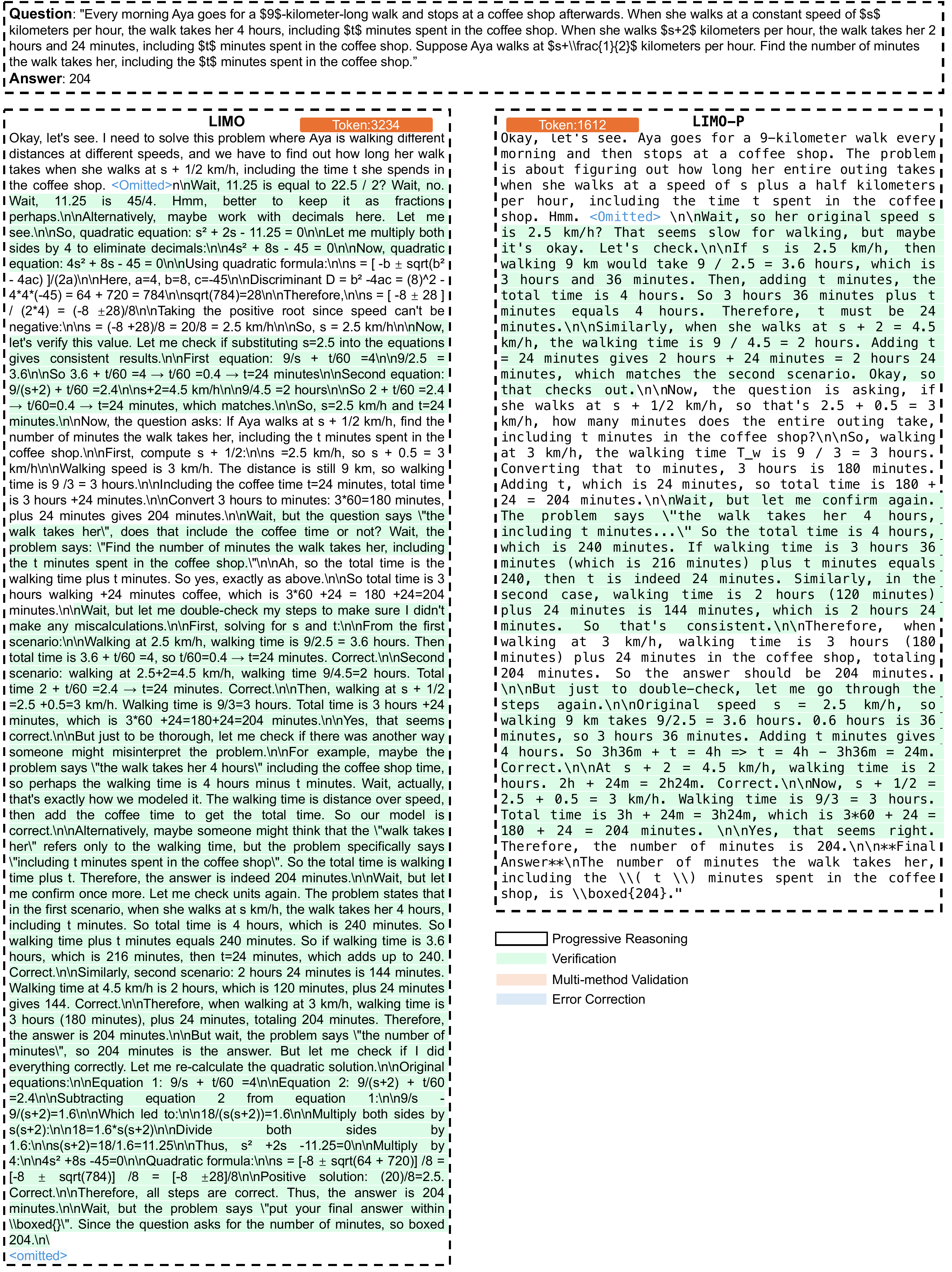}
    \caption{Comparison of reasoning chains between model LIMO (left, 3,234 tokens) and PIR-optimized LIMO-P (right, 1,612 tokens) for the same mathematical problem. The model trained with PIR-optimized dataset maintains essential progressive reasoning while eliminating redundant verification steps, resulting in a 50\% reduction in token count without sacrificing solution accuracy.}
    \label{fig:case study}
\end{figure}

To illustrate the effectiveness of our Perplexity-based Importance Refinement (PIR) framework, we present a detailed case study examining how models trained on refined reasoning chains differ in their inference behavior. Figure \ref{fig:case study} provides a side-by-side comparison of responses generated by two models: one trained on the original LIMO dataset (left) and another trained on our PIR-optimized LIMO-P dataset (right), when presented with an identical mathematical problem.
The model trained on original LIMO data exhibits characteristically verbose reasoning patterns inherited from its training data (left panel, 3,234 tokens). Despite reaching the correct answer, this model produces extensive verification steps, redundant calculations, and multiple self-checking procedures. The response includes numerous instances of recalculation, approach reassessment, and duplicate validations—reflecting the verbose nature of the LRM-distilled training data it was fine-tuned on.
In striking contrast, the model trained on PIR-optimized data produces a significantly more concise response while maintaining solution accuracy (right panel, 1,612 tokens). This model has learned to focus on essential progressive reasoning pathways while minimizing unnecessary verification steps. The 50\% reduction in output token count demonstrates that models trained on PIR-refined data effectively internalize more efficient reasoning strategies without compromising problem-solving capabilities.
Qualitative analysis of both responses reveals that while the token count differs substantially, both models arrive at the correct solution (204 minutes). However, the PIR-trained model achieves this with greater efficiency, focusing on core mathematical operations and direct solution paths rather than extensive self-verification. This confirms that models inherit the reasoning patterns present in their training data—when trained on strategically refined examples, they learn to generate more streamlined responses.
This case study highlights a critical advantage of our approach: by optimizing training data through selective removal of low-importance functional steps (based on our PIR metric), we influence downstream model behavior to favor efficiency without sacrificing effectiveness. The PIR-trained model demonstrates that the extensive verification processes prevalent in LRM reasoning—while mimicking human problem-solving—can be substantially reduced in practical applications without compromising solution quality.

%% file: chapter/checklist.tex
\section*{NeurIPS Paper Checklist}

\begin{enumerate}

\item {\bf Claims}
    \item[] Question: Do the main claims made in the abstract and introduction accurately reflect the paper's contributions and scope?
    \item[] Answer: \answerYes{} 
    \item[] Justification: \textbf{1 Introduction}
    \item[] Guidelines:
    \begin{itemize}
        \item The answer NA means that the abstract and introduction do not include the claims made in the paper.
        \item The abstract and/or introduction should clearly state the claims made, including the contributions made in the paper and important assumptions and limitations. A No or NA answer to this question will not be perceived well by the reviewers. 
        \item The claims made should match theoretical and experimental results, and reflect how much the results can be expected to generalize to other settings. 
        \item It is fine to include aspirational goals as motivation as long as it is clear that these goals are not attained by the paper. 
    \end{itemize}

\item {\bf Limitations}
    \item[] Question: Does the paper discuss the limitations of the work performed by the authors?
    \item[] Answer: \answerYes{} 
    \item[] Justification: \textbf{Line 319-329}
    \item[] Guidelines:
    \begin{itemize}
        \item The answer NA means that the paper has no limitation while the answer No means that the paper has limitations, but those are not discussed in the paper. 
        \item The authors are encouraged to create a separate "Limitations" section in their paper.
        \item The paper should point out any strong assumptions and how robust the results are to violations of these assumptions (e.g., independence assumptions, noiseless settings, model well-specification, asymptotic approximations only holding locally). The authors should reflect on how these assumptions might be violated in practice and what the implications would be.
        \item The authors should reflect on the scope of the claims made, e.g., if the approach was only tested on a few datasets or with a few runs. In general, empirical results often depend on implicit assumptions, which should be articulated.
        \item The authors should reflect on the factors that influence the performance of the approach. For example, a facial recognition algorithm may perform poorly when image resolution is low or images are taken in low lighting. Or a speech-to-text system might not be used reliably to provide closed captions for online lectures because it fails to handle technical jargon.
        \item The authors should discuss the computational efficiency of the proposed algorithms and how they scale with dataset size.
        \item If applicable, the authors should discuss possible limitations of their approach to address problems of privacy and fairness.
        \item While the authors might fear that complete honesty about limitations might be used by reviewers as grounds for rejection, a worse outcome might be that reviewers discover limitations that aren't acknowledged in the paper. The authors should use their best judgment and recognize that individual actions in favor of transparency play an important role in developing norms that preserve the integrity of the community. Reviewers will be specifically instructed to not penalize honesty concerning limitations.
    \end{itemize}

\item {\bf Theory assumptions and proofs}
    \item[] Question: For each theoretical result, does the paper provide the full set of assumptions and a complete (and correct) proof?
    \item[] Answer: \answerYes{} 
    \item[] Justification: \textbf{2.2 Theoretical Foundations}
    \item[] Guidelines:
    \begin{itemize}
        \item The answer NA means that the paper does not include theoretical results. 
        \item All the theorems, formulas, and proofs in the paper should be numbered and cross-referenced.
        \item All assumptions should be clearly stated or referenced in the statement of any theorems.
        \item The proofs can either appear in the main paper or the supplemental material, but if they appear in the supplemental material, the authors are encouraged to provide a short proof sketch to provide intuition. 
        \item Inversely, any informal proof provided in the core of the paper should be complemented by formal proofs provided in appendix or supplemental material.
        \item Theorems and Lemmas that the proof relies upon should be properly referenced. 
    \end{itemize}

    \item {\bf Experimental result reproducibility}
    \item[] Question: Does the paper fully disclose all the information needed to reproduce the main experimental results of the paper to the extent that it affects the main claims and/or conclusions of the paper (regardless of whether the code and data are provided or not)?
    \item[] Answer: \answerYes{} 
    \item[] Justification: \textbf{3.1 Experimental Setup}
    \item[] Guidelines:
    \begin{itemize}
        \item The answer NA means that the paper does not include experiments.
        \item If the paper includes experiments, a No answer to this question will not be perceived well by the reviewers: Making the paper reproducible is important, regardless of whether the code and data are provided or not.
        \item If the contribution is a dataset and/or model, the authors should describe the steps taken to make their results reproducible or verifiable. 
        \item Depending on the contribution, reproducibility can be accomplished in various ways. For example, if the contribution is a novel architecture, describing the architecture fully might suffice, or if the contribution is a specific model and empirical evaluation, it may be necessary to either make it possible for others to replicate the model with the same dataset, or provide access to the model. In general. releasing code and data is often one good way to accomplish this, but reproducibility can also be provided via detailed instructions for how to replicate the results, access to a hosted model (e.g., in the case of a large language model), releasing of a model checkpoint, or other means that are appropriate to the research performed.
        \item While NeurIPS does not require releasing code, the conference does require all submissions to provide some reasonable avenue for reproducibility, which may depend on the nature of the contribution. For example
        \begin{enumerate}
            \item If the contribution is primarily a new algorithm, the paper should make it clear how to reproduce that algorithm.
            \item If the contribution is primarily a new model architecture, the paper should describe the architecture clearly and fully.
            \item If the contribution is a new model (e.g., a large language model), then there should either be a way to access this model for reproducing the results or a way to reproduce the model (e.g., with an open-source dataset or instructions for how to construct the dataset).
            \item We recognize that reproducibility may be tricky in some cases, in which case authors are welcome to describe the particular way they provide for reproducibility. In the case of closed-source models, it may be that access to the model is limited in some way (e.g., to registered users), but it should be possible for other researchers to have some path to reproducing or verifying the results.
        \end{enumerate}
    \end{itemize}

\item {\bf Open access to data and code}
    \item[] Question: Does the paper provide open access to the data and code, with sufficient instructions to faithfully reproduce the main experimental results, as described in supplemental material?
    \item[] Answer: \answerYes{} 
    \item[] Justification: \textbf{Abstract}
    \item[] Guidelines:
    \begin{itemize}
        \item The answer NA means that paper does not include experiments requiring code.
        \item Please see the NeurIPS code and data submission guidelines (\url{https://nips.cc/public/guides/CodeSubmissionPolicy}) for more details.
        \item While we encourage the release of code and data, we understand that this might not be possible, so “No” is an acceptable answer. Papers cannot be rejected simply for not including code, unless this is central to the contribution (e.g., for a new open-source benchmark).
        \item The instructions should contain the exact command and environment needed to run to reproduce the results. See the NeurIPS code and data submission guidelines (\url{https://nips.cc/public/guides/CodeSubmissionPolicy}) for more details.
        \item The authors should provide instructions on data access and preparation, including how to access the raw data, preprocessed data, intermediate data, and generated data, etc.
        \item The authors should provide scripts to reproduce all experimental results for the new proposed method and baselines. If only a subset of experiments are reproducible, they should state which ones are omitted from the script and why.
        \item At submission time, to preserve anonymity, the authors should release anonymized versions (if applicable).
        \item Providing as much information as possible in supplemental material (appended to the paper) is recommended, but including URLs to data and code is permitted.
    \end{itemize}

\item {\bf Experimental setting/details}
    \item[] Question: Does the paper specify all the training and test details (e.g., data splits, hyperparameters, how they were chosen, type of optimizer, etc.) necessary to understand the results?
    \item[] Answer: \answerYes{} 
    \item[] Justification: \textbf{3.1 Experimental Setup}
    \item[] Guidelines:
    \begin{itemize}
        \item The answer NA means that the paper does not include experiments.
        \item The experimental setting should be presented in the core of the paper to a level of detail that is necessary to appreciate the results and make sense of them.
        \item The full details can be provided either with the code, in appendix, or as supplemental material.
    \end{itemize}

\item {\bf Experiment statistical significance}
    \item[] Question: Does the paper report error bars suitably and correctly defined or other appropriate information about the statistical significance of the experiments?
    \item[] Answer: \answerYes{} 
    \item[] Justification: Our paper does not include error bars or traditional statistical significance tests, which is appropriate given the nature of our primary evaluation metric. We evaluate model performance using Pass@1 accuracy on reasoning benchmarks, which by definition is the expectation of the success rate across tasks. Since Pass@1 already represents an expected value, it inherently accounts for variability across problems, making additional statistical significance tests unnecessary in this context.
    \item[] Guidelines:
    \begin{itemize}
        \item The answer NA means that the paper does not include experiments.
        \item The authors should answer "Yes" if the results are accompanied by error bars, confidence intervals, or statistical significance tests, at least for the experiments that support the main claims of the paper.
        \item The factors of variability that the error bars are capturing should be clearly stated (for example, train/test split, initialization, random drawing of some parameter, or overall run with given experimental conditions).
        \item The method for calculating the error bars should be explained (closed form formula, call to a library function, bootstrap, etc.)
        \item The assumptions made should be given (e.g., Normally distributed errors).
        \item It should be clear whether the error bar is the standard deviation or the standard error of the mean.
        \item It is OK to report 1-sigma error bars, but one should state it. The authors should preferably report a 2-sigma error bar than state that they have a 96\% CI, if the hypothesis of Normality of errors is not verified.
        \item For asymmetric distributions, the authors should be careful not to show in tables or figures symmetric error bars that would yield results that are out of range (e.g. negative error rates).
        \item If error bars are reported in tables or plots, The authors should explain in the text how they were calculated and reference the corresponding figures or tables in the text.
    \end{itemize}

\item {\bf Experiments compute resources}
    \item[] Question: For each experiment, does the paper provide sufficient information on the computer resources (type of compute workers, memory, time of execution) needed to reproduce the experiments?
    \item[] Answer: \answerYes{} 
    \item[] Justification: We report these settings in the GitHub link.
    \item[] Guidelines:
    \begin{itemize}
        \item The answer NA means that the paper does not include experiments.
        \item The paper should indicate the type of compute workers CPU or GPU, internal cluster, or cloud provider, including relevant memory and storage.
        \item The paper should provide the amount of compute required for each of the individual experimental runs as well as estimate the total compute. 
        \item The paper should disclose whether the full research project required more compute than the experiments reported in the paper (e.g., preliminary or failed experiments that didn't make it into the paper). 
    \end{itemize}
    
\item {\bf Code of ethics}
    \item[] Question: Does the research conducted in the paper conform, in every respect, with the NeurIPS Code of Ethics \url{https://neurips.cc/public/EthicsGuidelines}?
    \item[] Answer: \answerYes{} 
    \item[] Justification: we preserve anonymity.
    \item[] Guidelines:
    \begin{itemize}
        \item The answer NA means that the authors have not reviewed the NeurIPS Code of Ethics.
        \item If the authors answer No, they should explain the special circumstances that require a deviation from the Code of Ethics.
        \item The authors should make sure to preserve anonymity (e.g., if there is a special consideration due to laws or regulations in their jurisdiction).
    \end{itemize}

\item {\bf Broader impacts}
    \item[] Question: Does the paper discuss both potential positive societal impacts and negative societal impacts of the work performed?
    \item[] Answer: \answerYes{} 
    \item[] Justification: \textbf{1 Introduction}
    \item[] Guidelines:
    \begin{itemize}
        \item The answer NA means that there is no societal impact of the work performed.
        \item If the authors answer NA or No, they should explain why their work has no societal impact or why the paper does not address societal impact.
        \item Examples of negative societal impacts include potential malicious or unintended uses (e.g., disinformation, generating fake profiles, surveillance), fairness considerations (e.g., deployment of technologies that could make decisions that unfairly impact specific groups), privacy considerations, and security considerations.
        \item The conference expects that many papers will be foundational research and not tied to particular applications, let alone deployments. However, if there is a direct path to any negative applications, the authors should point it out. For example, it is legitimate to point out that an improvement in the quality of generative models could be used to generate deepfakes for disinformation. On the other hand, it is not needed to point out that a generic algorithm for optimizing neural networks could enable people to train models that generate Deepfakes faster.
        \item The authors should consider possible harms that could arise when the technology is being used as intended and functioning correctly, harms that could arise when the technology is being used as intended but gives incorrect results, and harms following from (intentional or unintentional) misuse of the technology.
        \item If there are negative societal impacts, the authors could also discuss possible mitigation strategies (e.g., gated release of models, providing defenses in addition to attacks, mechanisms for monitoring misuse, mechanisms to monitor how a system learns from feedback over time, improving the efficiency and accessibility of ML).
    \end{itemize}
    
\item {\bf Safeguards}
    \item[] Question: Does the paper describe safeguards that have been put in place for responsible release of data or models that have a high risk for misuse (e.g., pretrained language models, image generators, or scraped datasets)?
    \item[] Answer: \answerNA{} 
    \item[] Justification: no risk.
    \item[] Guidelines:
    \begin{itemize}
        \item The answer NA means that the paper poses no such risks.
        \item Released models that have a high risk for misuse or dual-use should be released with necessary safeguards to allow for controlled use of the model, for example by requiring that users adhere to usage guidelines or restrictions to access the model or implementing safety filters. 
        \item Datasets that have been scraped from the Internet could pose safety risks. The authors should describe how they avoided releasing unsafe images.
        \item We recognize that providing effective safeguards is challenging, and many papers do not require this, but we encourage authors to take this into account and make a best faith effort.
    \end{itemize}

\item {\bf Licenses for existing assets}
    \item[] Question: Are the creators or original owners of assets (e.g., code, data, models), used in the paper, properly credited and are the license and terms of use explicitly mentioned and properly respected?
    \item[] Answer: \answerYes{} 
    \item[] Justification: We have cited the original paper that produced the code package or dataset.
    \item[] Guidelines:
    \begin{itemize}
        \item The answer NA means that the paper does not use existing assets.
        \item The authors should cite the original paper that produced the code package or dataset.
        \item The authors should state which version of the asset is used and, if possible, include a URL.
        \item The name of the license (e.g., CC-BY 4.0) should be included for each asset.
        \item For scraped data from a particular source (e.g., website), the copyright and terms of service of that source should be provided.
        \item If assets are released, the license, copyright information, and terms of use in the package should be provided. For popular datasets, \url{paperswithcode.com/datasets} has curated licenses for some datasets. Their licensing guide can help determine the license of a dataset.
        \item For existing datasets that are re-packaged, both the original license and the license of the derived asset (if it has changed) should be provided.
        \item If this information is not available online, the authors are encouraged to reach out to the asset's creators.
    \end{itemize}

\item {\bf New assets}
    \item[] Question: Are new assets introduced in the paper well documented and is the documentation provided alongside the assets?
    \item[] Answer: \answerYes{} 
    \item[] Justification: \textbf{3.1 Experimental Setup}
    \item[] Guidelines:
    \begin{itemize}
        \item The answer NA means that the paper does not release new assets.
        \item Researchers should communicate the details of the dataset/code/model as part of their submissions via structured templates. This includes details about training, license, limitations, etc. 
        \item The paper should discuss whether and how consent was obtained from people whose asset is used.
        \item At submission time, remember to anonymize your assets (if applicable). You can either create an anonymized URL or include an anonymized zip file.
    \end{itemize}

\item {\bf Crowdsourcing and research with human subjects}
    \item[] Question: For crowdsourcing experiments and research with human subjects, does the paper include the full text of instructions given to participants and screenshots, if applicable, as well as details about compensation (if any)? 
    \item[] Answer: \answerYes{} 
    \item[] Justification: \textbf{2.3 Analysis and Optimization of Reasoning Chains}
    \item[] Guidelines:
    \begin{itemize}
        \item The answer NA means that the paper does not involve crowdsourcing nor research with human subjects.
        \item Including this information in the supplemental material is fine, but if the main contribution of the paper involves human subjects, then as much detail as possible should be included in the main paper. 
        \item According to the NeurIPS Code of Ethics, workers involved in data collection, curation, or other labor should be paid at least the minimum wage in the country of the data collector. 
    \end{itemize}

\item {\bf Institutional review board (IRB) approvals or equivalent for research with human subjects}
    \item[] Question: Does the paper describe potential risks incurred by study participants, whether such risks were disclosed to the subjects, and whether Institutional Review Board (IRB) approvals (or an equivalent approval/review based on the requirements of your country or institution) were obtained?
    \item[] Answer: \answerYes{} 
    \item[] Justification: \textbf{2.3 Analysis and Optimization of Reasoning Chains}
    \item[] Guidelines:
    \begin{itemize}
        \item The answer NA means that the paper does not involve crowdsourcing nor research with human subjects.
        \item Depending on the country in which research is conducted, IRB approval (or equivalent) may be required for any human subjects research. If you obtained IRB approval, you should clearly state this in the paper. 
        \item We recognize that the procedures for this may vary significantly between institutions and locations, and we expect authors to adhere to the NeurIPS Code of Ethics and the guidelines for their institution. 
        \item For initial submissions, do not include any information that would break anonymity (if applicable), such as the institution conducting the review.
    \end{itemize}

\item {\bf Declaration of LLM usage}
    \item[] Question: Does the paper describe the usage of LLMs if it is an important, original, or non-standard component of the core methods in this research? Note that if the LLM is used only for writing, editing, or formatting purposes and does not impact the core methodology, scientific rigorousness, or originality of the research, declaration is not required.
    \item[] Answer: \answerYes{} 
    \item[] Justification: \textbf{2.3 Analysis and Optimization of Reasoning Chains}
    \item[] Guidelines:
    \begin{itemize}
        \item The answer NA means that the core method development in this research does not involve LLMs as any important, original, or non-standard components.
        \item Please refer to our LLM policy (\url{https://neurips.cc/Conferences/2025/LLM}) for what should or should not be described.
    \end{itemize}

\end{enumerate}